# A Heuristic Algorithm for the Fabric Spreading and Cutting Problem in Apparel Factories

Xiuqin Shang, Dayong Shen, Fei-Yue Wang, Fellow, IEEE, and Timo R. Nyberg

*Abstract*—We study the fabric spreading and cutting problem in apparel factories. For the sake of saving the material costs, the cutting requirement should be met exactly without producing additional garment components. For reducing the production costs, the number of lays that corresponds to the frequency of using the cutting beds should be minimized. We propose an iterated greedy algorithm for solving the fabric spreading and cutting problem. This algorithm contains a constructive procedure and an improving loop. Firstly the constructive procedure creates a set of lays in sequence, and then the improving loop tries to pick each lay from the lay set and rearrange the remaining lays into a smaller lay set. The improving loop will run until it cannot obtain any small lay set or the time limit is due. The experiment results on 500 cases shows that the proposed algorithm is effective and efficient.

## I. INTRODUCTION

Cutting and packing problems (Dyckhoff & Finke, 1992) are classic combinatorial optimization problems that address the optimal utilization of resources. Cutting problems regard the best use of materials such as cloth, paper, wood, and steel. The efficient use of material contributes to the economical utilization of natural resources. It is also of great economic importance in production processes.

In this paper we consider the fabric spreading and cutting problem (FSCP in short) in apparel factories which is a special case of cutting problems. Given the production plan of producing a certain number of garments of various styles and sizes, FSCP consists of spreading a set of available fabrics on a cutting bed and cutting them into garment components. The consumption of fabrics as well as the use of the cutting bed should be minimized (Jacobs-Blecha et al. 1998, Rose and Shier 2007, Nascimento et al. 2010).

FSCP is an NP-complete combinatorial problem for the solution space rises exponentially with the number of required garment styles and sizes, and fabric types and colors (Jacobs-Blecha et al. 1998). Existing algorithms mainly focus on middle and small instances (Wong W K 2003, Nascimento et al. 2010). For large-scale apparel factories often receive large orders in a production cycle, we will focus on the middle and large instances in this paper.

Section 1 introduces FSCP. Section 2 brings out a brief review on FSCP. Section 3 describes the mathematical model for FSCP and proposes a heuristic algorithm for solving it. Section 4 evaluates the proposed algorithm. Section 5 concludes the paper.

## II. LITERATURE REVIEW

Cutting and packing problems form an old and very well-known family, called CP (Cutting & Packing) in Dyckhoff (1990), Paul and Elizabeth (1992) and Wäscher et al. (2007). CP has received growing attention in numerous real-world applications such as computer science, industrial engineering, logistics, manufacturing, and so on in recent years. As a branch of CP, FSCP has not yet obtained sufficient attention in academic research although it plays an important role in industrial application.

Some early publications focus on minimizing the time for fabric spreading and cutting in the garment industry. Wong et al. (2000) present a spreading and cutting sequencing model using GA to solve the sequencing problem of the spreading and cutting system. Wong, W K (2003) proposes a generic optimized table-planning model to optimize apparel manufacturing resource allocation. To minimize the sum of the cutting cost and fabric cost, Jacobs-Blecha et al. (1998) suggest a method to solve the cut order planning problem for apparel manufacturing. They consider the cutting order that only contains garments of the same fabric type and color, and can be completed in a lay with layers of different lengths.

Degraeve and Vandebroek (1998) formulate the problem as a mixed integer programming problem where the cutting setup costs and excess production are examined. They assume that all templates (referred to as stencils) have the same length independent of the garment size. All cutting pattern types should be enumerated in the proposed method. Degraeve et al. (2002) extend this work by using a non-linear integer programming solver.

Rose and Shier (2007) suggest a tree search method for cut scheduling in the apparel industry. By this method, the number of lays is minimized while no excess garment components are produced. They assume that the requirement contains only garments with the same fabric type and color, and all templates have the same length independent of garment size. Nascimento et al. (2010) propose a state-space approach to solve the fabric spreading and cutting problem. By assuming a fixed number of lays, the objective is to minimize the comprehensive cost of the fabric spreading and cutting process.

The number of cutting patterns will grow exponentially when the number of garment style and size (or the quantity of fabric type and color) increases. The above methods are effective and efficient for solving small scale fabric spreading and cutting problem. But they will suffer the situation called state explosion and become powerless when solving middle and large scale problems that exist in many factories.

Toscano, et al. (2017) discuss the two dimensional cutting stock problem that occurs in small-scale furniture factories. They bring out a heuristic approach to minimize both the number of stocks and the number of saw cycles. Hereby a saw cycle in a furniture factory is similar to a lay in a garment factory. The cutting stock problem with minimizing the number of stocks in a furniture factory can be considered as a three dimensional cutting stock problem with some constraints. If all the small items cannot rotate and have the same height as the one of the stock, this problem will be equivalent to the fabric spreading and cutting problem with single fabric type and color.

Vanzela et al. (2017) discuss the integrated lot sizing and cutting stock problem with saw cycle constraints in furniture factories. An effective method is brought out to reduce raw material waste as well as item (referred as pieces) production and inventory costs. The above two publications only consider stocks of the same type. In this paper the fabric spreading and cutting problem includes garments of various types and colors. Thus we will study the fabric spreading and cutting problem in apparel factories and bring out a new algorithm.

Wuttke and Heese (2018) study the two-dimensional cutting stock problem with sequence dependent setup times. They formulate the sequencing problem as a mixed integer program (MIP) and derive a near-optimal algorithm. Foerster and Wäscher (2000) bring out a heuristic approach for solving the one-dimensional cutting stock problem with pattern reduction. The approach firstly creates a cutting plan then reduces the number of patterns used in this plan. Yanasse and Limeira (2006) solve the cutting stock problem with pattern reduction in three steps. Patterns that will be used in the cutting plan are created in the first step. The complete cutting plan is obtained in the second step. The pattern combination method in Foerster and Wäscher (2000) is used in the third step to reduce the pattern number.

Cui et al. (2015) analyze the one-dimensional cutting stock problem with pattern reduction and solve it in two steps. A set of patterns are created in the first step, an integer linear programming (ILP) model is solved to minimize the sum of material and setup costs over the given pattern set in the second step.Cui et al. (2014) present a heuristic algorithm for solving the two-dimensional arbitrary stock-size cutting stock problem. Ma et al. (2018) formulate the combined cutting stock and lot-sizing problem with pattern setup as a mixed-integer linear programming model (MILP) and provide a dynamic programming-based heuristic (DPH) to solve it. Cui and Zhao (2013) present a heuristic approach to solve the rectangular two-dimensional single stock size cutting stock problem with two-staged patterns. In this approach the column-generation method is repeatedly applied to create patterns until all small items are produced. Alvarez-Valdés et al. (2007) combine greedy randomized adaptive search procedure (GRASP) and path relink (PR) into an algorithm to solve the two-dimensional two-staged cutting stock problem.

Some efficient approaches are proposed to solve the constrained two-staged two-dimensional cutting problem. These methods include the exact algorithm based on the bottom-up strategy by Hifi and M'Hallah (2005), approximate algorithms based on strip generation by Hifi and M'Hallah (2006), and the approach that combines strip-generation procedure with beam search by Hifi et al. (2008). The proposed Algorithms performances continue to be improved.

We will design an iterated greedy algorithm for solving FSCP. The above methods are beneficial references for some procedures such as dynamic programming and sequential heuristic are applied in our algorithm [25-29].

III. MATHEMATICAL MODEL AND HEURISTIC ALGORITHM

An apparel factory usually produces a number of different garment styles in a production cycle (e.g., week). Each of these styles consists of distinct couples of fabric types and garment sizes. Each of these couples is called a stock keeping unit (SKU) that corresponds to a certain garment style with the specific fabric type and garment size.

The apparel manufacturing process involves three main steps, namely fabric spreading and cutting, garment sewing, and garment finishing. On the fabric spreading and cutting phase, the fabric rolls of different types and colors are spread over a cutting bed in many layers and a knife (or laser) follows a predetermined route to cut the fabrics into garment components. These components are then tied and put forward to the garment sewing step. The multiple layers of fabric on a cutting bed for a cut are called a lay.

The detailed introduction to fabric spreading and cutting can be in Nascimento et al. (2010). As shown in Fig. 1, each size of each garment style has a template that contains all of its garment components. All components of a template are positioned in a rectangular area in a way of ensuring minimum loss of fabric. The template of a specific garment style and size require a fixed length of fabric. Templates are placed in a line to form a cutting pattern. Each lay is cut according to a cutting pattern. All SKUs that are cut simultaneously according to a template in a lay are called a column.

Manufacturing costs in the fabric spreading and cutting stage account for a great proportion of the whole manufacturing costs. Manufacturing costs in this stage consist of the costs of using cutting machines, the costs of the waste cloth and etc. The costs of using cutting machines rely on the lay number while the costs of the waste cloth arise from the overcut garment components. For the sake of reducing the whole manufacturing costs, the scheduling of fabric spreading and cutting should be improved.

Given the requirements of all SKUs in a production cycle, we try to generate a cutting plan with minimized lay number while all required garment components are cut without cutting any extra ones. Firstly, some necessary definitions are made in Table I.

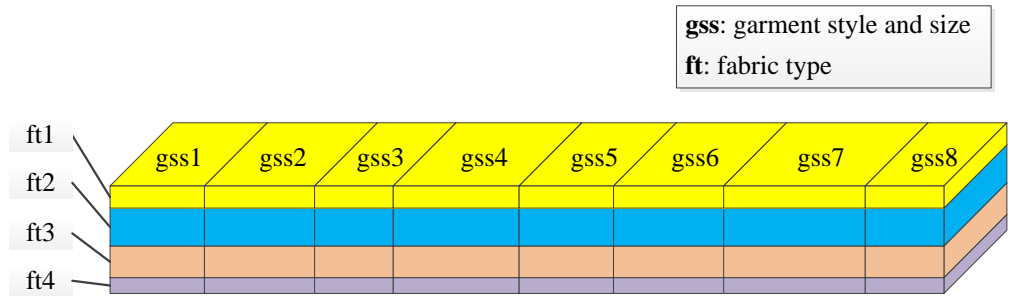

Fig. 1 A lay pattern for fabric spreading and cutting

**Table I. The definitions for FSCP**

| name | meaning |
|---|---|
| $l\_ub$ | the maximum allowed fabric length of the cutting bed |
| $h\_ub$ | the maximum allowed fabric layer number of the cutting bed |
| $g$ | the number of different garment figures, which is the product of the number of garment styles and the number of sizes of each garment style |
| $f$ | the number of different fabric types |
| $s_{ij} (i=1,2,\cdots,g; j=1,2,\cdots,f)$ | the requirement of the SKU with the $i$th garment figure and the $j$th fabric type |
| $l_i (i=1,2,\cdots,g)$ | the required fabric length of the $i$th garment figure |
| $S = \begin{bmatrix} s_{11} & s_{12} & \cdots & s_{1f} \\ s_{21} & s_{22} & \cdots & s_{2f} \\ \vdots & \vdots & \ddots & \vdots \\ s_{g1} & s_{g2} & \cdots & s_{gf} \end{bmatrix}$ | the matrix that stores the requirements of all the SKUs |
| $L = (l_1, l_2, \cdots, l_g)$ | the vector that stores the required fabric lengths of all the garment figures |
| $a$ | the number of required lays |
| $q_{ki} (k=1,2,\cdots,a; i=1,2,\cdots,g)$ | the number of the $i$th garment figure on the cutting pattern of the $k$th lay |
| $h_{kj} (k=1,2,\cdots,a; j=1,2,\cdots,f)$ | the layer number of the $j$th fabric type in the $k$th lay |
| $V_k, k=1,2,\cdots,a$ | the volume of the $k$th lay |
| $UR_k, k=1,2,\cdots,a$ | the volume and utilization rate of the $k$th lay |

With the above variables, FSCP can be illustrated as:

$$\min a \tag{1}$$

subject to:

$$\sum_{k=1}^{a} q_{ki} h_{kj} = s_{ij}; i=1,2,\cdots,g; j=1,2,\cdots,f \quad (2)$$

$$\sum_{i=1}^{g} l_i q_{ki} \leq l\_ub; \quad k=1,2,\cdots,a \quad (3)$$

$$\sum_{j=1}^{f} h_{kj} \leq h\_ub; \quad k=1,2,\cdots,a \quad (4)$$

$$V_k = \sum_{i=1}^{g} l_i q_{ki} * \sum_{j=1}^{f} h_{kj}; k=1,2,\cdots,a \quad (5)$$

$$UR_k = V_k / (l\_ub * h\_ub); k=1,2,\cdots,a \quad (6)$$

Formula (1) is the objective function that minimizes the number of total used lays. Formula (2) ensures that the production requirement is satisfied exactly. Formula (3) and Formula (4) guarantee that the length and height of each lay do not exceed the maximum allowed length and the maximum allowed height, respectively. Formula (5) computes the volume of the $k$th lay. Formula (6) can obtain the volume and utilization rate of the $k$th lay.

The proposed algorithm in this paper is referred to as HFSC (**H**euristic algorithm for **F**abric **S**preading and **C**utting). HFSC includes a constructive procedure to create an initial solution and an improving loop to refine the solution.

The main procedure of HFSC is illustrated in Procedure 1. The lay set *BEST_LAY* denotes the ultimate solution. Firstly an initial solution *LAY* is obtained by invoking CreateLays (described in Procedure 2). Then each lay in *LAY* is examined whether it can be transferred to *BEST_LAY*. If the remaining lays in *LAY* can be rearranged into less number of new lays after a lay is taken away from *LAY*, the lay will be transferred to *BEST_LAY* and *LAY* is replaced with the new lays. If no such lay exists in *LAY*, all lays in *LAY* will be added to *BEST_LAY*.

Procedure 1.

HFSC$(S, L, l\_ub, h\_ub)$

   $BEST\_LAY := \varnothing$

   $LAY := \{lay_1, lay_2, \cdots\} := \text{Construction}(S, L, l\_ub, h\_ub)$

   do

     for each lay $lay_k$ in *LAY*

       Let $S'$ store the numbers of SKUs included in *LAY* except $lay_k$

       $LAY' := \{lay'_1, lay'_2, \cdots\} := \text{Construction}(S', L, l\_ub, h\_ub)$

       if $(|LAY'|+1 < |LAY|)$

         Add $lay_k$ to *BEST_LAY*

         $LAY := LAY'$

         break

   while $(|BEST\_LAY| \text{ changes})$

   $BEST\_LAY := BEST\_LAY \cup LAY$

   return *BEST_LAY*

As shown in Procedure 2, Construction () creates and constructs a set of lays in series according to the input parameters. In order to accelerate the overall algorithm, the variables *ref_v* and *ref_h* are introduced with $l\_ub*h\_ub$ and $h\_ub$ as their initial values, respectively. Construction (described in Procedure 3) is invoked repeatedly to create new lays until all required SKUs are produced. Once a new lay is created, the values of *ref_v* and *ref_h* will be replaced with the mean volume and height of all created lays, respectively.

Procedure 2.

CreateLays$(S, L, l\_ub, h\_ub)$

   $k := 1$

   $ref\_v := l\_ub * h\_ub$

   $ref\_h := h\_ub$

   while ($S$ contains non-zero elements)

     $lay_k := \text{Construction}(S, L, ref\_v, ref\_h, l\_ub)$

     Remove SKUs in $lay_k$ from $S$

     $ref\_v :=$ the mean volume of $lay_1, lay_2, \cdots, lay_k$

     $ref\_h :=$ the mean height of $lay_1, lay_2, \cdots, lay_k$

     $k := k+1$

   return $\{lay_1, lay_2, \cdots, lay_k\}$

Construction () is illustrated in Procedure 3. Firstly, all possible heights for each fabric type and color are created by invoking CreatePossibleHeights (displayed in Procedure 4). For each composition of heights of different fabric types and colors that add up to *ref_h*, a set of columns is created by invoking CreateColumns (displayed in Procedure 5). With these columns, a lay is created by solving the one-dimensional knapsack model:

$$KS\left(l\_ub, \{(l_1, qc_1),(l_2, qc_2),\cdots,(l_n, qc_n)\}\right)$$

$$= \max\left\{\sum_{i=1}^{g} l_i * qu_i \middle| \begin{array}{l} \{(l_1, qu_1),(l_2, qu_2),\cdots,(l_n, qu_n)\} \text{ is a feasible solution;} \\ 0 \leq qu_i \leq qc_i, (i=1,2,\cdots,n); \\ \sum_{i=1}^{g} l_i * qu_i \leq l\_ub \end{array}\right\}$$

(7)

For the sake of saving computation time, Construction () will return the best found lay once it find a lay whose volume exceeds *ref_v* or it cannot find better lays any more.

Procedure 3.

CreateLay$(S, L, ref\_v, ref\_h, l\_ub)$

$\left[\{ch_{11}, ch_{12}, \cdots\}, \{ch_{21}, ch_{22}, \cdots\}, \cdots, \{ch_{f1}, ch_{f2}, \cdots\}\right]$
$:= $ CreatePossibleHeights$(S, L, l\_ub, ref\_h)$

Let *best_lay* be the best lay so far

while $\begin{pmatrix} \text{the volume of } best\_lay < ref\_v \text{ and} \\ ref\_h * l\_B > \text{the volume of } best\_lay \end{pmatrix}$

for each $[h_1, h_2, \cdots, h_g]$ where $h_j \in \{ch_{j1}, ch_{j2}, \cdots\}$ $(j=1,2,\cdots,f)$ and $\sum_{j=1}^{f} h_j = ref\_h$

$\{c_1, c_2, \cdots\} := $ CreateColumns$(S, L, [h_1, h_2, \cdots, h_f])$

$new\_lay := $ KS$(l\_ub, \{c_1, c_2, \cdots\})$

if (the volume of $new\_lay$ > the volume of $best\_lay$)
  $best\_lay := new\_lay$
  if (the volume of $best\_lay \geq ref\_v$)   break

$ref\_h := ref\_h - 1$

return *best_lay*

As illustrated in Procedure 4, CreatePossibleHeights() creates an integer set for each fabric type and color. Each number in the integer set for the *j*th ($1 \leq j \leq f$) fabric type corresponds to a possible layer number of the *j*th ($1 \leq j \leq f$) fabric type in the will-be-created lay. The possible layer number of the *j*th ($1 \leq j \leq f$) fabric type in the lay is among *ref_h*, 0 and the divisors of the required numbers of the SKUs with the *j*th ($1 \leq j \leq f$) fabric type. These divisors must be less than *ref_h* and guarantee the corresponding SKUs could be completely produced in the lay.

Procedure 4.

CreatePossibleHeights$(S, L, l\_ub, ref\_h)$

for each $j \in \{1, 2, \cdots, f\}$
  $CH_j := \{ \}$
  if $\left(\exists s_{ij} \geq ref\_h, i \in \{1, 2, \cdots, g\}\right)$
    add *ref_h* to $CH_j$
  for each $ph \in \{ref\_h - 1, ref\_h - 2, \cdots, 1\}$
    if $\left(\exists s_{ij} \neq 0, \text{and } \left[\frac{s_{ij}}{ph}\right] = \frac{s_{ij}}{ph}, \text{and } \frac{s_{ij}}{ph} * l_i \leq l\_ub, i \in \{1, 2, \cdots, g\}\right)$
      add *ph* to $CH_j$
  add 0 to $CH_j$
return $\{CH_1, CH_2, \cdots, CH_f\}$

The procedure CreateColumns() creates columns for the SKUs of each garment style and size according to the layer number of each fabric type and color in the will-be-created lay. $[h_1, h_2, \cdots, h_f]$ denotes the layer number of each fabric type and color in this lay.

⋯Procedure 5.

CreateColumns$\left(R, L, [h_1, h_2, \cdots, h_f]\right)$

for each $i \in \{1, 2, \cdots, g\}$
  $qc_i := \min\left\{\frac{s_{ij}}{h_j}, j \in \{1, 2, \cdots, f\} \text{ and } h_j \neq 0\right\}$
  $column_i := (l_i, qc_i)$
return $\{column_1, column_2, \cdots, column_g\}$

## IV. EXPERIMENTAL EVALUATION

The proposed HFSC algorithm is implemented and compared with SS-HR proposed by reference [12] in C#, and run on a server with Intel Core2 Duo Q9400@2.66GHz and Microsoft Windows 7 Ultimate. The compiling environment is Microsoft Visual Studio 2012.

We design ten groups (named G1, G2, … , G10) of test cases for evaluating HFSC. Each group contains 50 cases. Each of the 500 cases contains the required numbers of SKUs of 5 garment styles. Each garment style contains 6 garment sizes and 5 fabric types and colors. For each case, *l_B*, *h_B* and *L* are 720, 160 and (60, 63, 66, 69, 73, 76, 69, 72, 75, 78, 82, 86, 80, 83, 86, 90, 94, 98, 90, 94, 98, 102, 106, 110, 99, 103, 107, 111, 115, 120) respectively. The required numbers of SKUs in the 500 cases are generated randomly using the random seed 1000000 in sequence of G1-G10. Table II lists the lower bounds (LBs) and the upper bounds (UBs) of the required numbers of SKUs in the 500 cases, respectively. The data files of the 500 cases are in the attachments.

**Table II. The LBs and the UBs of the required numbers of SKUs in the cases of G1-G10**

| | LB | UB |
|---|---|---|
| G1 | 300 | 400  |
| G2 | 300 | 600  |
| G3 | 400 | 500  |
| G4 | 300 | 800  |
| G5 | 400 | 700  |
| G6 | 500 | 600  |
| G7 | 300 | 1000 |
| G8 | 400 | 900  |
| G9 | 500 | 800  |
| G10| 600 | 700  |

Table III lists the results comparison between SS-HR and HFSC on the ten groups of 500 test cases. In the experiment, $K$ denotes the ultimate solution of the number of the cutting bed. Accordingly, $UR$ denotes the utilization rate of the ultimate solution. Besides, $TC$ is the time (a second as a unit) that the computer program spends in solving a case. To prevent HFSC from spending too much time on a case, the limit of the running time of HFSC for a case is 1200 seconds.

Table III shows that HFSC improves the mean $K$ and the mean $UR$ compared with SS-HR on G1-G10, whiles the mean cutting bed number K of HFSC is less than one of SS-HR, and the utilization rate of cutting bed of HFSC increases compared with SS-HR. For the cases with the same LB and UB of the required numbers of SKUs, when the difference of the required numbers of SKUs reduces, the mean $UR$ raises slightly. The detailed results of the G1-G10 are listed Table IV – Table VII, respectively.

Table IV shows that HFSC may obtain a solution ranging from 62 to 66 for each of the 50 cases in G1 in a reasonable time. The difference between the maximum solution and the minimum one is about 6%.

**Table III. The result comparisons between HFSC and SS-HR on G1 – G10**

|     | Mean $K$ | | Mean $UR$ (%) | |
|-----|-------|-------|-------|-------|
|     | SS-HR | HFSC  | SS-HR | HFSC  |
| G1  | 66.04 | **63.92** | 61.20 | **63.19** |
| G2  | 75.22 | **73.14** | 68.77 | **70.72** |
| G3  | 75.16 | **73.06** | 69.08 | **71.06** |
| G4  | 86.04 | **83.4**  | 73.69 | **76.01** |
| G5  | 85.70 | **82.78** | 73.97 | **76.57** |
| G6  | 85.00 | **82.5**  | 74.50 | **76.75** |
| G7  | 96.16 | **93.36** | 77.53 | **79.85** |
| G8  | 96.66 | **93.74** | 77.90 | **80.31** |
| G9  | 95.98 | **93.3**  | 78.04 | **80.27** |
| G10 | 95.88 | **93.2**  | 78.09 | **80.37** |

**Table IV. The detailed results of HFSC on G1**

| case | $K$ | case | $K$ | case | $K$ | case | $K$ | case | $K$ |
|------|----|------|----|------|----|------|----|------|----|
| 1 | 65 | 11 | 63 | 21 | 63 | 31 | 65 | 41 | 64 |
| 2 | 65 | 12 | 64 | 22 | 64 | 32 | 63 | 42 | 63 |
| 3 | 65 | 13 | 64 | 23 | 62 | 33 | 64 | 43 | 63 |
| 4 | 63 | 14 | 64 | 24 | 64 | 34 | 64 | 44 | 65 |
| 5 | 63 | 15 | 64 | 25 | 65 | 35 | 66 | 45 | 62 |
| 6 | 65 | 16 | 64 | 26 | 63 | 36 | 63 | 46 | 64 |
| 7 | 63 | 17 | 62 | 27 | 63 | 37 | 65 | 47 | 64 |
| 8 | 64 | 18 | 64 | 28 | 64 | 38 | 64 | 48 | 64 |
| 9 | 65 | 19 | 65 | 29 | 66 | 39 | 64 | 49 | 64 |
| 10 | 65 | 20 | 64 | 30 | 64 | 40 | 63 | 50 | 63 |

The mean value of SKU numbers of the cases in G2 is approximately same as the one in G3. Table V shows that HFSC may obtain a solution ranging from 71 to 75 for each of the 100 cases in G2 and G3 in a reasonable time. The minimum of the solutions for the 100 cases is about 5% less than the maximum one.

**Table V. The detailed results of HFSC on G2 and G3**

| G2 | | | | G3 | | | |
|---|---|---|---|---|---|---|---|
| case | $K$ | case | $K$ | case | $K$ | case | $K$ |
| 1 | 72 | 26 | 75 | 1 | 73 | 26 | 73 |
| 2 | 73 | 27 | 73 | 2 | 73 | 27 | 73 |
| 3 | 74 | 28 | 73 | 3 | 73 | 28 | 73 |
| 4 | 72 | 29 | 73 | 4 | 73 | 29 | 73 |
| 5 | 75 | 30 | 74 | 5 | 74 | 30 | 72 |
| 6 | 74 | 31 | 74 | 6 | 73 | 31 | 73 |
| 7 | 73 | 32 | 72 | 7 | 74 | 32 | 72 |
| 8 | 74 | 33 | 73 | 8 | 72 | 33 | 73 |
| 9 | 73 | 34 | 73 | 9 | 73 | 34 | 73 |
| 10 | 73 | 35 | 72 | 10 | 73 | 35 | 74 |
| 11 | 73 | 36 | 73 | 11 | 73 | 36 | 74 |
| 12 | 72 | 37 | 74 | 12 | 73 | 37 | 73 |
| 13 | 73 | 38 | 74 | 13 | 74 | 38 | 73 |
| 14 | 74 | 39 | 74 | 14 | 73 | 39 | 73 |
| 15 | 72 | 40 | 73 | 15 | 72 | 40 | 73 |
| 16 | 72 | 41 | 72 | 16 | 73 | 41 | 73 |
| 17 | 75 | 42 | 73 | 17 | 74 | 42 | 74 |
| 18 | 71 | 43 | 72 | 18 | 73 | 43 | 73 |
| 19 | 72 | 44 | 74 | 19 | 72 | 44 | 73 |
| 20 | 73 | 45 | 73 | 20 | 73 | 45 | 73 |
| 21 | 73 | 46 | 73 | 21 | 73 | 46 | 73 |
| 22 | 75 | 47 | 74 | 22 | 73 | 47 | 73 |
| 23 | 73 | 48 | 74 | 23 | 73 | 48 | 73 |
| 24 | 73 | 49 | 74 | 24 | 73 | 49 | 73 |
| 25 | 73 | 50 | 71 | 25 | 74 | 50 | 73 |

Table VI shows that the cases in G4, G5 and G6 approximately have the same mean SKU number. HFSC may obtain a solution ranging from 81 to 86 for each of the 150 cases in G4, G5 and G6 in a reasonable time. The maximum solution is about 6% greater than the minimum one.

For the most massive 200 cases in G7-G10, HFSC may obtain a solution ranging from 89 to 96 for each of them in a reasonable time, , as shown in Table VII. The difference between the maximum solution and the minimum one is about 7%.

In the experiments, the results show that HFSC is superior to SS-HR through the detailed results mean *K* and mean *UR* in the 500 cases. Therefore, the conclusion can be obtained that the proposed method is efficient and has a good performance.

**TableVI. The detailed results of HFSC on G4 – G6**

| | G4 | | | | G5 | | | | G6 | | |
|---|---|---|---|---|---|---|---|---|---|---|---|
| case | *K* | case | *K* | case | *K* | case | *K* | case | *K* | case | *K* |
| 1 | 84 | 26 | 83 | 1 | 84 | 26 | 82 | 1 | 82 | 26 | 83 |
| 2 | 82 | 27 | 83 | 2 | 83 | 27 | 83 | 2 | 83 | 27 | 81 |
| 3 | 84 | 28 | 85 | 3 | 83 | 28 | 83 | 3 | 83 | 28 | 82 |
| 4 | 83 | 29 | 85 | 4 | 82 | 29 | 82 | 4 | 84 | 29 | 82 |
| 5 | 83 | 30 | 82 | 5 | 84 | 30 | 83 | 5 | 82 | 30 | 82 |
| 6 | 82 | 31 | 84 | 6 | 83 | 31 | 81 | 6 | 83 | 31 | 82 |
| 7 | 81 | 32 | 82 | 7 | 83 | 32 | 82 | 7 | 83 | 32 | 82 |
| 8 | 86 | 33 | 82 | 8 | 81 | 33 | 83 | 8 | 83 | 33 | 83 |
| 9 | 83 | 34 | 82 | 9 | 82 | 34 | 82 | 9 | 82 | 34 | 82 |
| 10 | 83 | 35 | 84 | 10 | 84 | 35 | 83 | 10 | 83 | 35 | 83 |
| 11 | 82 | 36 | 83 | 11 | 84 | 36 | 83 | 11 | 83 | 36 | 82 |
| 12 | 84 | 37 | 85 | 12 | 84 | 37 | 85 | 12 | 82 | 37 | 83 |
| 13 | 84 | 38 | 85 | 13 | 81 | 38 | 83 | 13 | 83 | 38 | 82 |
| 14 | 82 | 39 | 85 | 14 | 84 | 39 | 84 | 14 | 83 | 39 | 83 |
| 15 | 83 | 40 | 84 | 15 | 84 | 40 | 83 | 15 | 83 | 40 | 83 |
| 16 | 84 | 41 | 82 | 16 | 85 | 41 | 80 | 16 | 83 | 41 | 82 |
| 17 | 83 | 42 | 84 | 17 | 81 | 42 | 83 | 17 | 81 | 42 | 83 |
| 18 | 82 | 43 | 83 | 18 | 83 | 43 | 83 | 18 | 84 | 43 | 83 |
| 19 | 84 | 44 | 82 | 19 | 81 | 44 | 83 | 19 | 82 | 44 | 83 |
| 20 | 84 | 45 | 85 | 20 | 83 | 45 | 84 | 20 | 81 | 45 | 82 |
| 21 | 84 | 46 | 82 | 21 | 82 | 46 | 83 | 21 | 82 | 46 | 83 |
| 22 | 84 | 47 | 84 | 22 | 82 | 47 | 83 | 22 | 82 | 47 | 83 |
| 23 | 84 | 48 | 83 | 23 | 83 | 48 | 82 | 23 | 83 | 48 | 82 |
| 24 | 81 | 49 | 86 | 24 | 82 | 49 | 81 | 24 | 83 | 49 | 82 |
| 25 | 83 | 50 | 86 | 25 | 84 | 50 | 83 | 25 | 82 | 50 | 82 |

**Table VII. The detailed results of HFSC on G7 – G10**

| | G7 | | | | G8 | | | | G9 | | | | G10 | | |
|---|---|---|---|---|---|---|---|---|---|---|---|---|---|---|---|
| case | *K* | case | *K* | case | *K* | case | *K* | case | *K* | case | *K* | case | *K* | case | *K* |
| 1 | 94 | 26 | 93 | 1 | 94 | 26 | 91 | 1 | 95 | 26 | 93 | 1 | 94 | 26 | 93 |
| 2 | 93 | 27 | 91 | 2 | 93 | 27 | 92 | 2 | 95 | 27 | 95 | 2 | 92 | 27 | 92 |
| 3 | 95 | 28 | 92 | 3 | 93 | 28 | 93 | 3 | 94 | 28 | 92 | 3 | 93 | 28 | 93 |
| 4 | 91 | 29 | 96 | 4 | 95 | 29 | 95 | 4 | 92 | 29 | 93 | 4 | 92 | 29 | 94 |
| 5 | 95 | 30 | 89 | 5 | 93 | 30 | 93 | 5 | 95 | 30 | 95 | 5 | 94 | 30 | 93 |
| 6 | 92 | 31 | 94 | 6 | 96 | 31 | 96 | 6 | 92 | 31 | 94 | 6 | 94 | 31 | 92 |
| 7 | 95 | 32 | 94 | 7 | 92 | 32 | 93 | 7 | 92 | 32 | 93 | 7 | 93 | 32 | 93 |
| 8 | 91 | 33 | 94 | 8 | 96 | 33 | 95 | 8 | 92 | 33 | 91 | 8 | 93 | 33 | 94 |
| 9 | 95 | 34 | 96 | 9 | 94 | 34 | 93 | 9 | 91 | 34 | 93 | 9 | 93 | 34 | 93 |
| 10 | 96 | 35 | 92 | 10 | 94 | 35 | 95 | 10 | 94 | 35 | 94 | 10 | 92 | 35 | 93 |
| 11 | 92 | 36 | 92 | 11 | 93 | 36 | 94 | 11 | 95 | 36 | 94 | 11 | 94 | 36 | 93 |
| 12 | 93 | 37 | 95 | 12 | 95 | 37 | 94 | 12 | 94 | 37 | 93 | 12 | 93 | 37 | 94 |
| 13 | 95 | 38 | 96 | 13 | 91 | 38 | 93 | 13 | 94 | 38 | 94 | 13 | 93 | 38 | 94 |
| 14 | 94 | 39 | 92 | 14 | 94 | 39 | 94 | 14 | 93 | 39 | 92 | 14 | 94 | 39 | 93 |
| 15 | 93 | 40 | 92 | 15 | 93 | 40 | 93 | 15 | 92 | 40 | 92 | 15 | 93 | 40 | 92 |
| 16 | 95 | 41 | 92 | 16 | 93 | 41 | 95 | 16 | 94 | 41 | 94 | 16 | 93 | 41 | 93 |
| 17 | 91 | 42 | 94 | 17 | 93 | 42 | 93 | 17 | 93 | 42 | 93 | 17 | 92 | 42 | 94 |
| 18 | 96 | 43 | 89 | 18 | 94 | 43 | 92 | 18 | 95 | 43 | 94 | 18 | 94 | 43 | 93 |
| 19 | 94 | 44 | 96 | 19 | 92 | 44 | 95 | 19 | 93 | 44 | 93 | 19 | 94 | 44 | 93 |
| 20 | 92 | 45 | 92 | 20 | 92 | 45 | 92 | 20 | 92 | 45 | 93 | 20 | 93 | 45 | 94 |
| 21 | 91 | 46 | 94 | 21 | 95 | 46 | 96 | 21 | 93 | 46 | 92 | 21 | 93 | 46 | 92 |
| 22 | 95 | 47 | 93 | 22 | 94 | 47 | 94 | 22 | 93 | 47 | 93 | 22 | 94 | 47 | 94 |
| 23 | 94 | 48 | 93 | 23 | 93 | 48 | 94 | 23 | 93 | 48 | 91 | 23 | 94 | 48 | 94 |
| 24 | 95 | 49 | 95 | 24 | 95 | 49 | 96 | 24 | 95 | 49 | 93 | 24 | 94 | 49 | 93 |
| 25 | 93 | 50 | 92 | 25 | 94 | 50 | 95 | 25 | 94 | 50 | 96 | 25 | 94 | 50 | 92 |

## V. CONCLUSIONS

This paper focuses on the fabric spreading and cutting problem in apparel industries. A heuristic algorithm is brought out to minimize the frequency of using the cutting beds while all required SKUs are produced without producing extra garment components. This algorithm includes a constructive procedure and an improving loop. The constructive procedure creates all lays in sequence according to given requirements. In the algorithm, firstly a lay set is obtained by invoking the constructive procedure, and then the improving loop tries to withdraw each lay from the lay set and invoke the constructive procedure to rearrange the left lays into a smaller lay set. The proposed algorithm is implemented in C#. A computational experiment that includes 500 cases is designed to test the computer program. For each case, the proposed algorithm gains effective and efficient result.